\documentclass[preprint,11pt,authoryear,a4paper]{elsarticle}
\usepackage[top=1in, bottom=1in, left=40mm, right=1in]{geometry}
\usepackage{setspace}
\usepackage{amssymb}
\usepackage[cmex10]{amsmath} 
\usepackage{multirow}
\usepackage{ulem}
\usepackage{algorithm}
\usepackage{algorithmic}
\usepackage{textcomp,mathcomp}

\usepackage{graphicx}
\graphicspath{{figure/}}
\usepackage{caption}
\usepackage{subcaption}
\usepackage{epstopdf}
\usepackage{xcolor}
\usepackage{hyperref}
\usepackage[super]{nth}
\hypersetup{colorlinks=true,bookmarksnumbered=true}
\newlength{\fwidth}

\usepackage[redeflists]{IEEEtrantools}
\usepackage{enumitem}

\usepackage{lmodern}

\newcommand{\XM}[1]{\ignorespaces}

\usepackage{enumitem}
\journal{International Journal of Forecasting}
\begin{document}
\sloppy
\hypersetup{urlcolor=black}
\begin{frontmatter}

\title{Predicting Peak Day and Peak Hour of Electricity Demand with Ensemble Machine Learning}

\author[]{Tao Fu}
\author[]{Huifen Zhou}
\author[]{Xu Ma}
\author[]{Z. Jason Hou}
\author[]{Di Wu\corref{cor1}}\ead{di.wu@pnnl.gov}

\address{Pacific Northwest National Laboratory, Richland, WA 99352 USA}
\cortext[cor1]{Corresponding author.}

\begin{abstract}
Battery energy storage systems can be used for peak demand reduction in power systems, leading to significant economic benefits.
Two practical challenges are 1) accurately determining the peak load days and hours and 2) quantifying and reducing uncertainties associated with the forecast in probabilistic risk measures for dispatch decision-making.
In this study, we develop a supervised machine learning approach to generate 1) the probability of the next operation day containing the peak hour of the month and 2) the probability of an hour to be the peak hour of the day. 
Guidance is provided on preparation and augmentation of data as well as selection of machine learning models and decision-making thresholds.
The proposed approach is applied to the Duke Energy Progress system and successfully captures 69 peak days out of 72 testing months with a 3\% exceedance probability threshold.
On 90\% of the peak days, the actual peak hour is among the 2 hours with the highest probabilities.
\end{abstract}

\end{frontmatter}

\newpage

\section{Introduction}
Many cooperatives, municipally owned utilities, and other types of load serving entities (LSE) purchase power from electricity markets or through power purchase contracts.
A capacity charge is paid based on the coincident demand during system peak hours.
Effectively reducing the peak demand leads to significant economic benefits.
Battery energy storage systems (BESS) are promising for peak demand reduction because of their flexibility and instantaneous response capability.
An LSE does not know exactly when the peak hour will occur.
Simply discharging BESS on all high-load days helps capture the peak hour and reduce the coincident demand, but causes unnecessary battery degradation and energy losses associated with charging/discharging.
In addition, due to limited energy capacity, BESS may not be able to discharge at the rated power in all high-load hours.
Therefore, advanced peak day and peak hour forecast methods are critical to maximizing benefits from BESS for peak demand reduction.

Predictions of the monthly peak hour can be derived from load forecast that spans the entire month.
Traditional methods, such as autoregressive integrated moving average (ARIMA), forecast load with univariate historical load.
More advanced nonlinear machine learning (ML) methods have also been proposed for load forecasting with multivariate predictors, including weather, calendar, and economics \citep{hong2016probabilistic}.
Examples of advanced nonlinear ML methods are
K-nearest neighbors (KNN) \citep{el2009forecasting},
fuzzy regression models \citep{hong2014fuzzy},
support vector machine (SVM) \citep{niu2010power},
gradient boosting machine (GBM) \citep{massaoudi2021novel},
random forest (RF) \citep{cheng2012random, huang2016permutation},
and artificial neural networks (ANN) \citep{el2000hybrid, ringwood2001forecasting, saini2008peak}.
These approaches can also be combined to improve load forecasting.
For example, \citet{el2000hybrid} developed a hybrid ANN with ARIMA for load forecasting.
\citet{el2009forecasting} proposed a multivariate load forecasting approach by combining support vector regression with a KNN local prediction framework.
A monthly peak hour can be identified based on the value of the hourly load forecast.
Such a method requires the hourly load forecast toward the end of the month.
There are significant uncertainties associated with long-term hourly load forecast.
A major challenge of predicting peak hours based on load forecast is how to model and quantify uncertainties, considering varying weather conditions (e.g., temperature, humidity, and wind speed) and nonlinear relationships between weather and load \citep{el2000hybrid}.

Efforts have also been made to directly predict peak hours.
For example, \citet{goodwin2016pattern} proposed an SVM and Gaussian mixture classification approach to directly estimate the peak hours for the next 7 days based on historical load.
Leveraging short-term load forecasts, \citet{jiang2014predicting} adopted a probabilistic approach for estimating the probability of the next day containing the highest hourly demand of a year.
A Naive Bayesian classification model was proposed for classifying whether an hour is a 5CP (top 5 coincident peaks of a year) hour \citep{ryu2016naive}.
\citet{liu2019effect} proposed a convolutional neural network (CNN) classification model to predict 24 hours ahead whether a day is a 5CP day.
For the peak hour model, they built separate models for summer and winter using different classification methods: Naive Bayes, SVM, RF, AdaBoost, CNN, long short-term memory (LSTM), and stacked autoencoder \citep{liu2019prediction}.
More recently, \citet{saxena2019hybrid} developed a hybrid classification model combining ARIMA, logistic regression (LR), and ANN for peak day prediction.

Both load forecast and direct peak prediction ML models require enough data for training and validation.
In practice, however, complete historical records of load and weather are not always available, which warrants the data augmentation effort.
In addition, these aforementioned approaches either ignored trailing or leading effects of factors or skipped dimension reduction and physical interpretation of the factor contributions.
To summarize, despite the progress to date, additional research and development are needed to better support BESS dispatch decision-making for peak demand reduction, including
1) quantifying uncertainties associated with peak day and peak hour predictions,
2) addressing the data inadequacy issue via ML data augmentation,
3) evaluating choices of peak day probability thresholds for decision-making considering both prediction accuracy and precision,
4) including all physical and temporal factors to fully capture their correlations and causalities with system load peaking behaviors, e.g., trailing and leading effects,
and 5) integrating ensemble ML model selection techniques to deal with factor collinearity, mixed data types, and overfitting, as well as trustworthy, explainable ML prediction.

In this paper, we propose an ensemble learning approach taking advantage of multiple ML techniques, including RF, GBM, and LR, for predicting peak day and peak hour to better support BESS dispatch decision-making.
To quantify the prediction uncertainties, we develop, validate, compare, and select optimal ML models to generate the probability of the next day to be the monthly peak day, and the probability of an hour to be the peak hour of a day.
The proposed approach also features ML-based data augmentation to address the data inadequacy issues and a procedure to select exceedance thresholds for decision-making.
We use a comprehensive set of predictors, including actual load in previous days of the month, day-ahead load forecast, weather forecast, and their derivatives, as well as temporal factors to help improve peak predictions.
By applying bagging and boosting techniques, the ensemble tree-based ML method can help avoid overfitting and deal with unbalanced data of mixed types, e.g., categorical vs. continuous \citep{domingos2012few, cieslak2008learning, breiman2001random, jaiswal2017application, friedman2001greedy}.
We also use naive load-based peak predictions to demonstrate the effectiveness of the proposed ML-based prediction framework.


\section{Dataset}\label{sec:dataset}
The proposed ML-based prediction framework is generic and can be applied to any system.
The Duke Energy Progress (DEP) system is used for illustration and analysis presented in this paper.
The input dataset includes historical load and weather data, which are described as follows.

\subsection{DEP load}
The DEP system consists of two balancing authorities: DEP East and DEP West.
The data are obtained from the Energy Information Administration \citep{EIA}, including both day-ahead forecasts and actual demand with an hourly resolution.
Complete records are only available for about 6 years starting from July 2015.
Data augmentation is proposed and used to generate 21 years of data for training and testing purposes, as described in Section~\ref{sec:augmentation}.

\begin{figure}[!t]
	\centering
	\includegraphics[width=0.7\columnwidth]{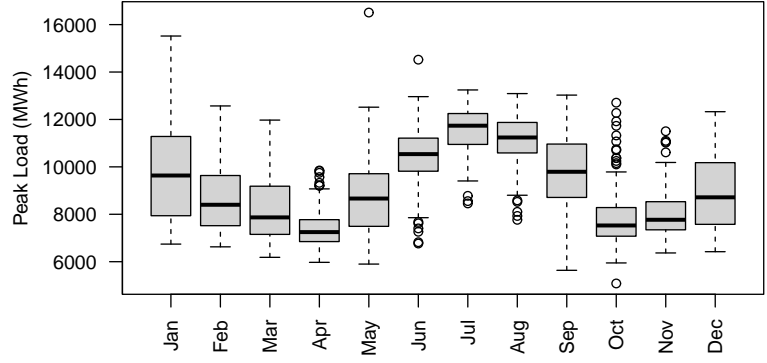}
	\caption{Boxplot of monthly peak load from July 2015 to December 2020.} \label{fig:boxplot_peak_load}
\end{figure}

\begin{figure}[!t]
	\centering
	\begin{subfigure}[b]{0.7\columnwidth}
		\includegraphics[width=1\linewidth]{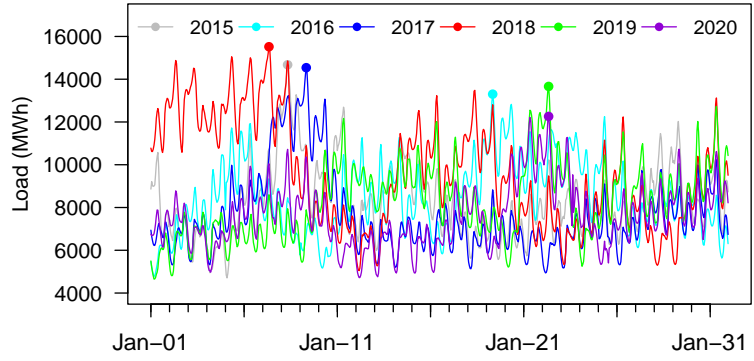}
		\caption{January}
	\end{subfigure}
	\begin{subfigure}[b]{0.7\columnwidth}
		\includegraphics[width=1\linewidth]{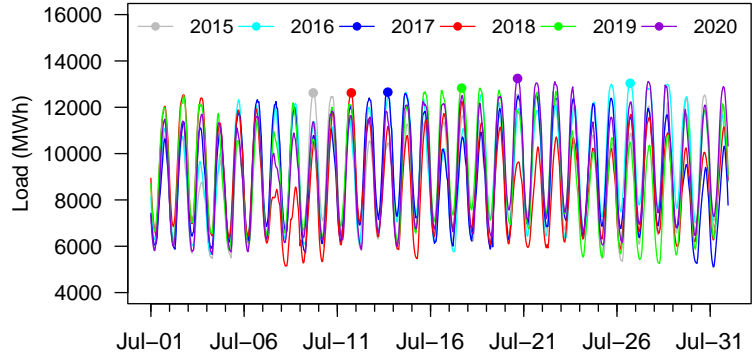}
		\caption{July}
	\end{subfigure}
	\caption{ Actual hourly load 2015-2020.} \label{fig:monthly_demand}
\end{figure}

A boxplot of monthly peak load is provided in Fig.~\ref{fig:boxplot_peak_load}.
As can be seen, the highest monthly peak loads typically occur in mid-winter (e.g., January) and mid-summer (e.g., July).  
Fig.~\ref{fig:monthly_demand} plots the system load vs. time for January and July.
The peak day can occur at the beginning, in the middle, or at the end of a month, and there is no obvious pattern, which increases the difficulty for peak day prediction.
The peak days in January exhibit strong variations over the years yet no dominant timings and durations of peak days. 

\begin{figure*}[!t]
	\centering
	\includegraphics[width=0.98\textwidth]{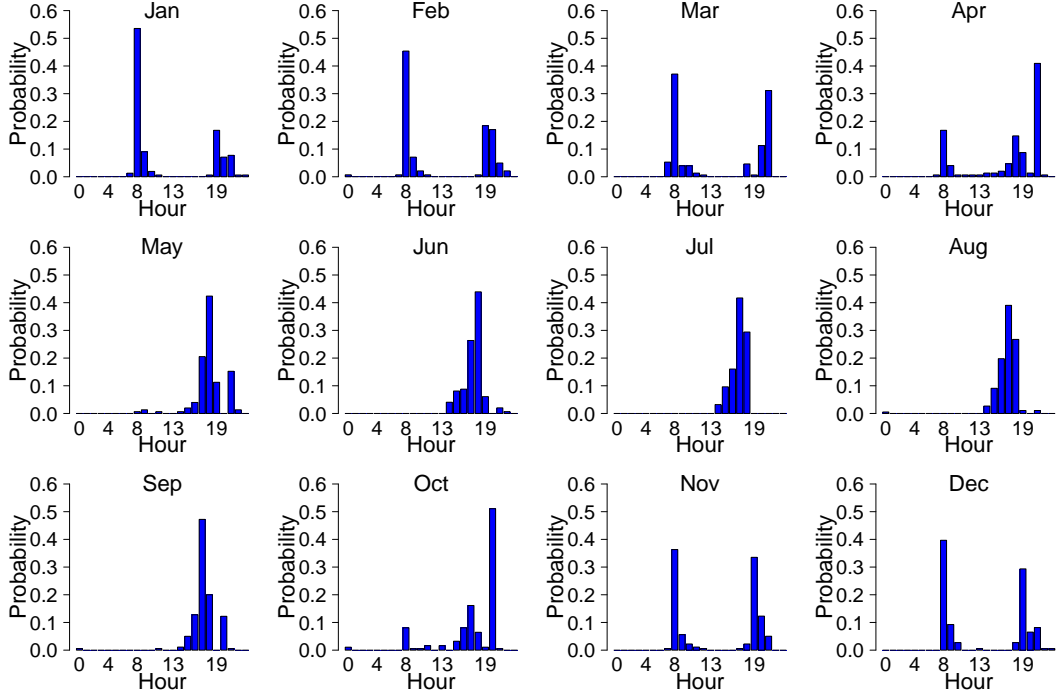}
	\caption{Distribution of daily peak hour by month from July 2015 to December 2020.} \label{fig:histogram_peak_hour}
\end{figure*}

Fig.~\ref{fig:histogram_peak_hour} shows the varying distributions of peak hours by month from July 2015 to December 2020.
In January and February, a majority of peak hours occurred at 8 a.m.
In March, November, and December, the peak hours had a bimodal distribution with one peak in the morning hours and the other peak in the evening hours.
In April and October, the peak hours happened in the later evening, hour 21 for April and hour 20 for October.
From May to September, a majority of peak hours were in the early evening hours, around hour 17 or 18.

\subsection{Weather data}
The weather datasets are obtained from the National Oceanic and Atmospheric Administration website \citep{noaa}.
In this study, the weather data at the Raleigh weather station are used as it is the closest weather station to the city of Raleigh, which is the geographic center of DEP and the second biggest city in North Carolina.
The raw weather data include dry-bulb air temperature, dew point temperature, wind speed, and visibility, with temporal resolutions between 15 minutes to an hour from 2000 to 2020.
To match the temporal resolution and coverage of the load data, the weather data are mapped to the nearest exact hours to generate hourly data.
In addition, the hourly humidity $H$ is calculated according to \citet{bosen1958approximation}:
\begin{equation}\label{eq:equ-humidity}
	H = \exp\left(\frac{17.625 \times DT}{243.04+DT}-\frac{17.625 \times T}{243.04+T}\right) \times 100\% \, \text{,}
\end{equation}
where $T$ is the dry-bulb air temperature (\textcelsius),
$DT$ is the dew point temperature (\textcelsius),
and $\exp(\cdot)$ is the exponential function.

\section{Methodology}\label{sec:methodology}

\subsection{Calculation of peak day and peak hour probabilities}\label{sec:peak_day_hour_calculation}
\subsubsection{Peak day model} \label{sec:peak_day_model}
\begin{figure}[!t]
	\centering	
	\begin{subfigure}{0.7\columnwidth}
		\includegraphics[width=1\linewidth]{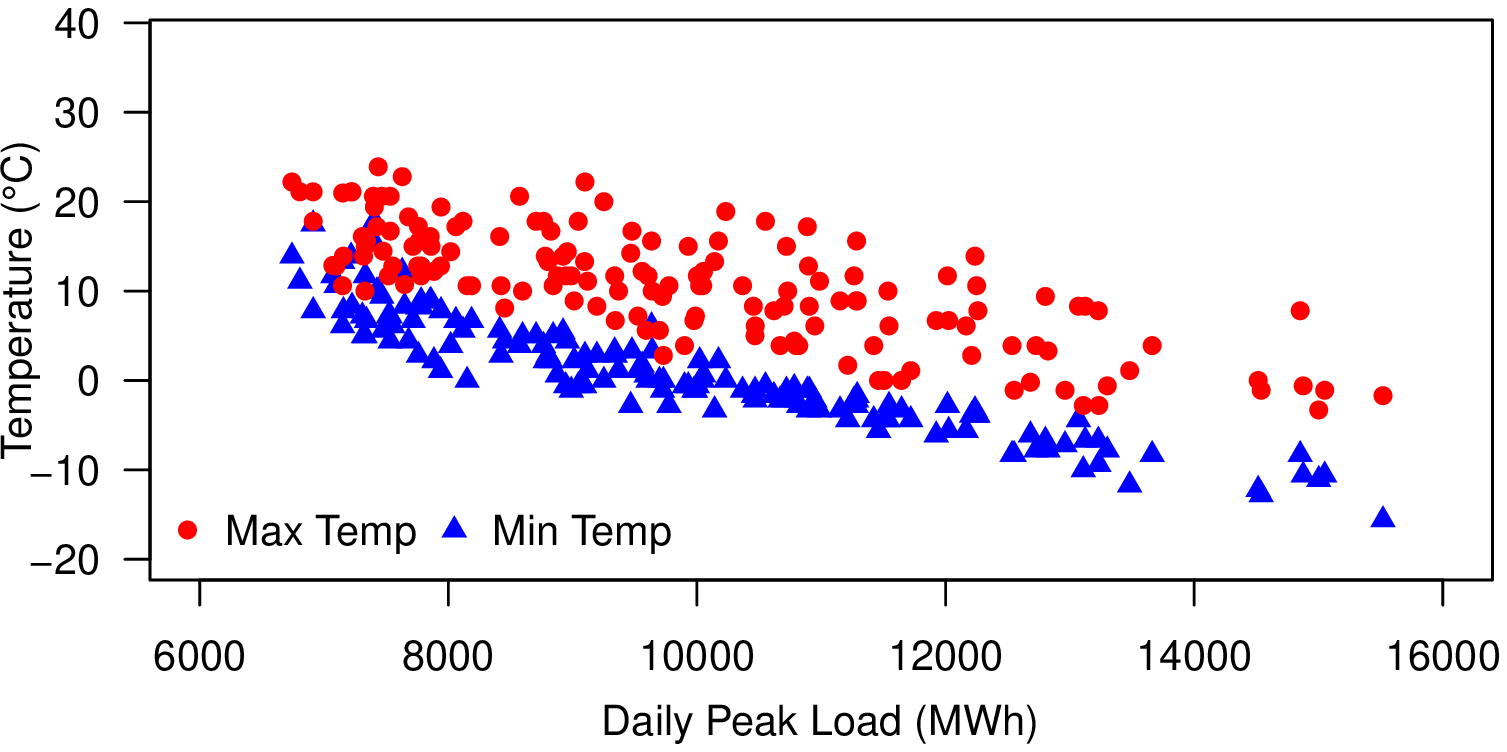}
		\caption{January} \label{fig:T_vs_Load_Jan}
	\end{subfigure}
	\begin{subfigure}{0.7\columnwidth}
		\includegraphics[width=1\linewidth]{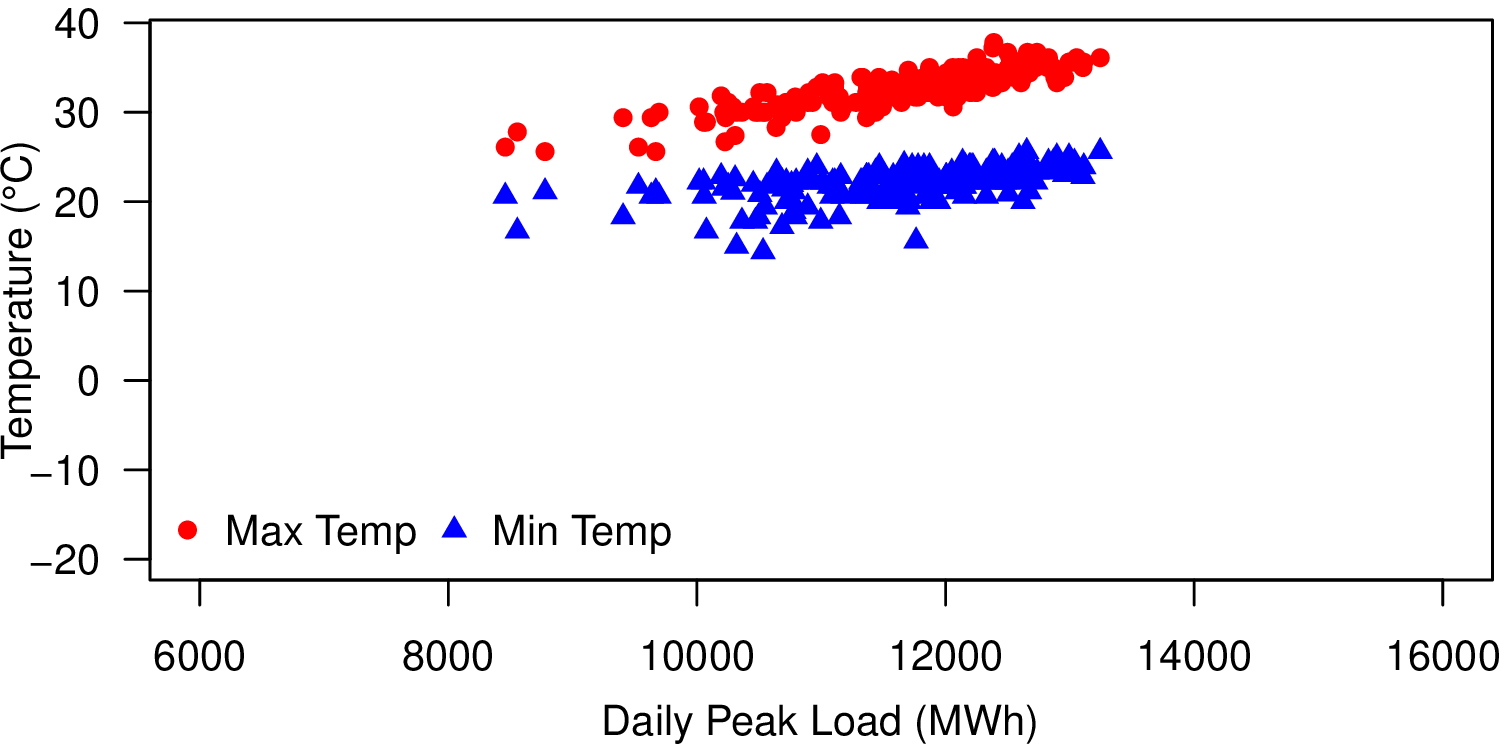}
		\caption{July} \label{fig:T_vs_Load_July}
	\end{subfigure}
	\caption{Daily peak load vs.\ daily maximum and minimum temperature.}
	\label{fig:T_vs_Load}
\end{figure}

\begin{table}[!t]
	\caption{R-squared of daily peak load vs.\ daily minimum and maximum temperature.}
	\label{R-squared}
	\setlength{\tabcolsep}{4pt}
	\centering
	\begin{tabular}{lllllllllllll}
		\hline
		& Jan. & Feb. & Mar. & Apr. & May  & June & July & Aug. & Sept. & Oct. & Nov. & Dec. \\
		\hline
		\texttt{T\_min} & 0.86 & 0.72 & 0.58 & 0.04 & 0.57 & 0.39 & 0.30 & 0.33 & 0.36  & 0.38 & 0.40 & 0.74 \\
		\hline
		\texttt{T\_max} & 0.58 & 0.50 & 0.29 & 0.11 & 0.80 & 0.80 & 0.73 & 0.77 & 0.77  & 0.50 & 0.26 & 0.56\\
		\hline
	\end{tabular}
\end{table}
There is a strong and nonlinear correspondence between power system load and weather  \citep{sobhani2020temperature}.
Fig.~\ref{fig:T_vs_Load} plots the DEP daily peak load vs. minimum and maximum temperature on each day in January and July from July 2015 to December 2020.
The R-squared values between the daily minimum/maximum temperatures and the daily peak load for all months are listed in Table~\ref{R-squared}.
Two key observations are highlighted as follows:
\begin{itemize}
	\item There is a strong negative correlation (with $R^{2} > 0.7$) between the daily minimum temperature and load in winter months (December to February),
	and a strong positive correlation between the daily maximum temperature and load in summer months (May to September).
	\item The correlation between daily peak load and daily minimum and maximum temperature is relatively weak ($R^{2} \leq 0.5$) in spring and fall (March, April, May, October, and November), which suggests that temperature is a less effective predictor of peak load during these months.
\end{itemize}

Based on these observations, both direct and indirect models are developed and tested in this study.
The selected predictors are the same for both models, including load and weather variables that are derived from the hourly data and a weekday/weekend indicator, as listed in Table~\ref{table-peak-day-predictors}.
Because of the lack of historical weather forecast data, actual weather data are used for the ML model development.
\begin{table}[!t]
	\caption{Predictors for the peak day model.} \label{table-peak-day-predictors}
	\centering
		\begin{tabular}{ll}
			\hline
			Predictor& Description \\
			\hline
			\texttt{load\_max}& Maximum load from the day-ahead forecast\\
			\texttt{T\_min\textsuperscript{*}}& Minimum temperature from the day-ahead forecast \\
			\texttt{T\_max\textsuperscript{*}}& Maximum temperature from the day-ahead forecast \\
			\texttt{weekdayIdx}& 0 or 1 indicating whether the next day is weekday or weekend\\
			\texttt{prev\_month\_max}& Maximum actual load of the month to date\\
			\texttt{prev\_MAX}& Maximum actual load on previous day\\
			\texttt{T\_min\_day\_2\_to\_7\textsuperscript{*}}& Minimum forecast temperature within the next 2 to 7 days\\
			\texttt{T\_max\_day\_2\_to\_7\textsuperscript{*}}& Maximum forecast temperature within the next 2 to 7 days\\
			\hline
	\end{tabular}
\end{table}
\begin{itemize}
  \item The direct model directly predicts whether the next day is the peak day of the month.
In this model, a binary response variable takes a value of 1 if a day is the monthly peak day and 0 otherwise.
Ideally, the predictors should also include the load and weather forecast toward the end of the month to better capture how future load may affect whether the next day is the peak day.
However, long-term load and weather forecasts are generally unavailable and therefore only forecasts of up to 7 days are used as predictors.
As a result, the direct model actually links partial month predictors to the full month peak day indicator.


\item The indirect model outputs the probability of the next day to be the up-to-date peak day, which is defined as the peak day of a time window starting from the beginning of the month and ending at day 7 into the future or the end of the month.
Therefore, the model links predictors to the peak day indicator at the exactly matched time window.
Then, the obtained up-to-date peak day probability ($P_\text{date}$) is converted to the monthly peak day probability ($P_\text{month}$) by applying a multiplier ($P_\text{mul}$) that reflects the chance of the monthly peak day occurring within the up-to-date time window, as expressed in \eqref{eq:equ-P_up_to_date}:
\begin{equation}\label{eq:equ-P_up_to_date}
	P_\text{month} = P_\text{date} \, P_\text{mul}\, .
\end{equation}
The multipliers corresponding to different time windows can be generated by examining the distribution of historical peak days.
It is found that the chance is generally proportional to the length of the time windows in a given month.
In other words, each day in a month has an equal probability to be the monthly peak day.
Therefore, the multiplier can be defined as
\begin{equation}
P_\text{mul} = \min [1, \frac{n+6}{N}]\, \text{,}
\end{equation}
where $N$ is the number of days of the current month and $n$ is the day of the month for the next day.
Unlike the direct model, which completely ignores the impacts of the load beyond 7 days, the indirect model takes future load into account in a stochastic manner.
\end{itemize}

\subsubsection{Peak hour model}
The peak hour model is trained using the hourly load and weather data.
The response variable is a binary variable that indicates whether an hour is the peak hour of the day.
The predictors include both hourly load forecast and weather data for the operating day, as well as a weekday or weekend binary indicator, as listed in Table~\ref{table-peak-hour-predictors}.
We have the following considerations for including the major predictors:
\begin{itemize}
  \item For each hour $t$, we use not only the temperature and humidity for the hour, as they directly impact the load components, e.g., heating and cooling, but also the temperature from $t-3$ to $t+3$ to explore the trailing or leading effects between temperature and load.
  \item Similarly, the forecast loads from hour $t-3$ to $t+3$ are included in the predictors because the peaks tend to cluster in groups. 
  \item The rank of load forecast for hour $t$ is included to distinguish the peak hour from other hours, particularly those with comparable loading levels.
  \item The maximums of load forecast before and after hour $t$ are included to describe the relative position of the load in hour $t$ with respect to highs in the past and future hours of the day.
  In comparison to load forecast of each hour beyond hours $t-3$ to $t+3$, the use of these two predictors helps reduce modeling complexity and avoid overfitting, while effectively capturing the overall impacts of the other hours on the operating day.
  \item In addition, as adjacent days tend to have more similar peaking behaviors, attributes from the previous day, including the rank of actual load for hour $t$, the forecast, and the actual load for hour $t$, are also used as predictors.
\end{itemize}

\begin{table}[!t]
	\caption{Predictors for the peak hour model.} \label{table-peak-hour-predictors}
	\centering
	\begin{tabular}{ll}
		\hline
		Predictor& 	Description \\
		\hline
		\texttt{load\_forecast} & Forecast load for hour $t$\\
		\texttt{T\textsuperscript{*}} & Forecast temperature for hour $t$ \\
		\texttt{humidity\textsuperscript{*}} & Forecast humidity at hour $t$ \\
		\texttt{weekendIdx} & A binary indicator: 1 if the operating day is on a weekend,\\
		& and 0 otherwise\\
		\texttt{peak\_prev\_day}& A binary indicator: 1 if hour $t$ is the peak hour on the\\
		& previous day, and 0 otherwise\\
		\texttt{T\_m\_1\textsuperscript{*}} & Forecast temperature for hour $t-1$ \\
		\texttt{T\_m\_2\textsuperscript{*}}& Forecast temperature for hour $t-2$\\
		\texttt{T\_m\_3\textsuperscript{*}}& Forecast temperature for hour $t-3$\\
		\texttt{T\_p\_1\textsuperscript{*}} & Forecast temperature for hour $t+1$ \\
		\texttt{T\_p\_2\textsuperscript{*}}& Forecast temperature for hour $t+2$\\
		\texttt{T\_p\_3\textsuperscript{*}}& Forecast temperature for hour $t+3$\\
		\texttt{load\_m\_1}& Forecast load for hour $t-1$\\
		\texttt{load\_m\_2}& Forecast load for hour $t-2$\\
		\texttt{load\_m\_3}& Forecast load for hour $t-3$ \\
		\texttt{load\_p\_1}& Forecast load for hour $t+1$ \\
		\texttt{load\_p\_2}& Forecast load for hour $t+2$\\
		\texttt{load\_p\_3}& Forecast load for hour $t+3$\\
		\texttt{prev\_max\_load} & Maximum forecast load from hour 0 to hour $t-1$\\
		\texttt{after\_max\_load} & Maximum forecast load from hour $t+1$ to 23\\
		\texttt{rank\_load\_forecast} & Rank of forecast load for hour $t$ on the operation day\\
		\texttt{rank\_load\_prevDay}& Rank of actual load for hour $t$ in on the previous day\\
		\texttt{load\_prevDay}& Actual load for hour $t$ on the previous day\\
		\texttt{load\_prevDay\_forecast}& Forecast load for hour $t$ on the previous day\\		
		\hline		
	\end{tabular}
\end{table}

\subsection{ML models}
Given the high dimensionality of the predictors and the expected non-linear relationships with response variables, we adopt the ensemble tree-based ML method: RF and GBM, also to avoid overfitting and deal with unbalanced data of mixed types, e.g., categorical vs. continuous.
RF is a tree-based ensemble learning method that can handle categorical variables, continuous variables, or a combination of both \citep{breiman2001random}.
RF constructs a number of ensemble trees, with each tree trained by a randomly selected subset of input data using a bootstrap aggregating technique.
At each node of a tree, instead of choosing the best split among all predictors, a randomly sampled subset of predictors is selected, and the best split is chosen from the subset predictors.
Each tree can grow to the maximum possible depth.
For classification problems, the final prediction is made by the majority votes from all the trees in the ensemble.
RF has been successfully applied to high-dimensionality systems.
Examples of its application in power systems include \citet{lahouar2015day,liu2021hybrid}.
RF models can be developed and optimized by refining key model configuration parameters such as the number of trees and tree depths, in addition to bootstrap re-sampling of subsets and multi-fold cross-validation.
A general RL-based classification algorithm is summarized in Algorithm \ref{alg:RF}.
In addition to estimating the probability of classification, an RF model also provides measures of relative feature importance of predictors.

\begin{algorithm}[!t]
	\caption{A General RF-based Classification Algorithm}
	\label{alg:RF}
	\begin{algorithmic}[1]
		\STATE Randomly split the total dataset into training and testing (e.g., 75\% and 25\%)
		\FOR{$i=1$ to \texttt{n\_tree}}
		\STATE Randomly select a subset $n$ out of $N$ samples
		\STATE Randomly select a subset of $m$ predictors out of $M$ factors (${x_1, x_2, \ldots, x_M}$)
		\STATE Grow a random forest tree \texttt{Ti} and prune it to the optimal depth \texttt{n\_depth}
		\STATE Output \texttt{Ti} tree forecast ${C_i(x_1, x_2, \ldots, x_M)}$, which is binary)
		\ENDFOR
		\STATE Generate RF prediction by the majority vote of all tree forecasts $C_i$, $\forall i=1, \ldots, \texttt{n\_tree}$
	\end{algorithmic}
\end{algorithm}

GBM is another tree-based ensemble learning method \citep{friedman2001greedy}.
Unlike RF, in which each tree is trained independently, GBM builds one tree at a time and the newly-built tree is added to previous trees to improve the overall prediction.
It can be viewed as an iterative numerical optimization process with a goal of finding an additive model that minimizes the loss function.
The new tree at each step is fitted to the residual of previous trees, and adding a new tree improves regions where previous trees did not perform well.
While the result of RF is voted by all the trees after the tree-building process, the result of GBM is optimized throughout the tree-building process.
Compared to RF, GBM can be more computationally intensive and more sensitive to noise in the training data set due to its iterative nature.

LR is a classical statistical ML method that models the relationship between a set of predictors and a categorical response variable using a logistic function (sigmoid function) \citep{stoltzfus2011logistic}.
The predictors of an LR model can be categorical and/or continuous variables, and the output is the probability of the response variable being one of the categorical outcomes.
Because there are no ``hyper-parameters'' that can be used to tune an LR model, it is ideally used as a baseline model in predictive analyses.

\subsection{Data augmentation} \label{sec:augmentation}
The raw load data are only available after July 2015, resulting in less than 80 monthly peak days.
Separate models by month are highly desirable to better capture varying load patterns in different months.
Therefore, data augmentation is needed to generate sufficient data for model training and reliable testing.
An ANN model for day-ahead load forecast has been developed using load and weather attributes\citep{berscheid2018open}.
The developed model has been cross-validated with historical load and weather data sets for major U.S. Balancing Authorities.
In this study, using the DEP weather and load data sets from July 2015 to December 2020, the developed ANN model was trained to predict hourly load (including both actual load and day-ahead forecast) with the following predictors: hour, temperatures for the current hour and the past 3 hours, humidity at the hour, day of the week for the predicting day, and weekday/weekend index.
After a hyperparameter search, an optimized ANN model containing two layers and 20 neurons was obtained to capture load patterns and the relationship between load and weather embedded in the raw load data.
Using the ANN model, we produced hourly actual load and day-ahead hourly load forecast over a 15-year period with actual weather data from 2000 to 2015.

To validate the ensemble ML model performance, a subset of 6-year data is randomly selected for model testing: 2001, 2006, 2008, 2011, 2019, and 2020.
The remaining 15-year data are used for training. 
Note that the actual weather data are used for the ML model development because of the lack of historical weather forecast data.

\section{Results}\label{sec:results}


\subsection{ML predictive model development}

\subsubsection{Model training}
\begin{figure}[!t]
	\centering
	\begin{subfigure}[b]{0.7\columnwidth}
		\includegraphics[width=1\textwidth]{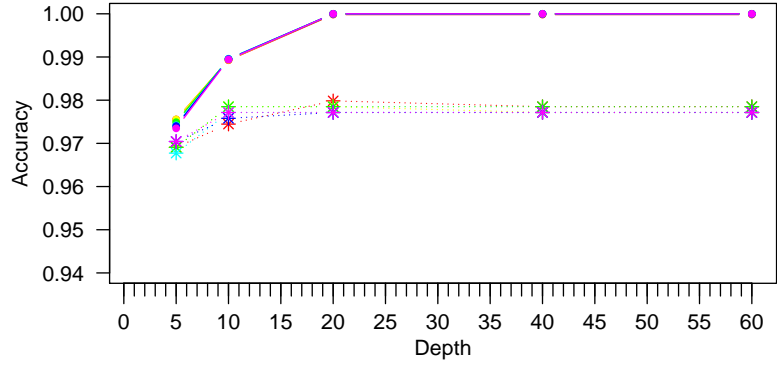}
		\caption{January}
	\end{subfigure}
	\begin{subfigure}[b]{0.7\columnwidth}
		\includegraphics[width=1\textwidth]{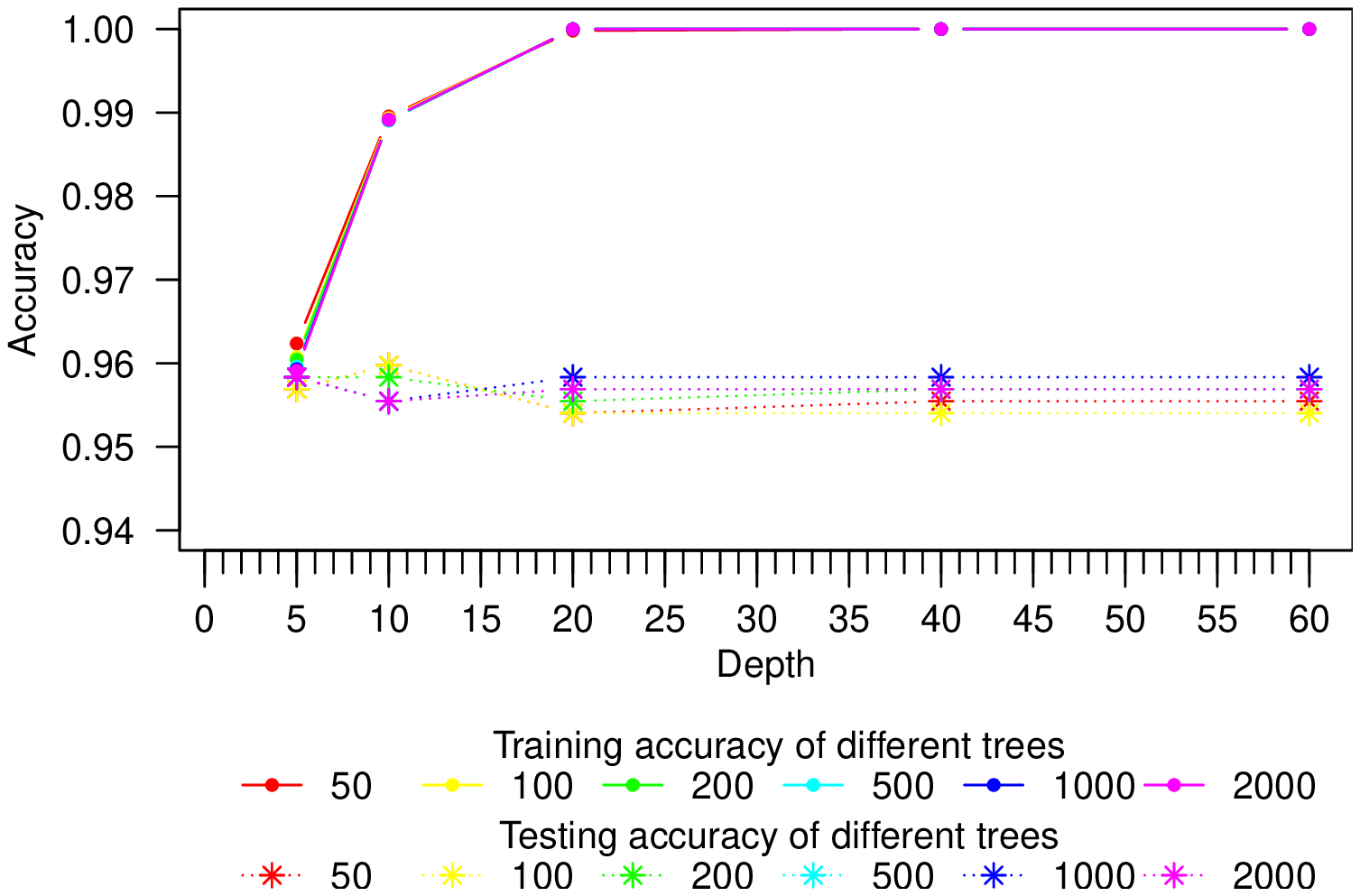}
		\caption{July}
	\end{subfigure}
	\caption{Peak hour RF model hyper-parameter tuning for January and July.} \label{fig:RF_hyper_parameter}
\end{figure}

Ensemble ML models, including LR, RF, and GBM, are developed for peak hour predictions.
For LR models, we use Akaike Information Criteria \citep{akaike1974new} for the model selection.
For RF and GBM models, we evaluate the testing accuracy with respect to two hyper-parameters: number of trees and tree depths.
Fig.~\ref{fig:RF_hyper_parameter} shows the training and testing accuracy with respect to different tree depths and different numbers of trees for both January and July RF models.
The selected  number of depths are 5, 10, 20, 40, and 60, and the selected number of trees are: 50, 100, 200, 500, 1000, and 2000.
The results show that the testing accuracy converges when the tree depths are greater than 20 for both models.
For the January model, the testing accuracy is comparable when the same number of depth is used regardless of the number of trees used.
For the July model, 1000 trees yield slightly higher accuracy than using other tree numbers when the tree depth is greater than 20.

\subsubsection{Model comparison}
For the peak hour models, each of the three ML ensembles is optimized with parameter search and cross-validation for reducing misfits and overfitting if possible.
The performance of the optimized final models is compared in terms of the overall accuracy of capturing actual peak hours for each month in the 6 testing years.
Overall, the RF models outperform both the LR and GBM models by 2\%,
In six monthly models --- January, April, June, July, August, and September --- the RF models perform better than the other two models by 5\% and 3\%, respectively.
LR models perform slightly better in February, May, October, and December; while GBM models perform better in March and November.
Because RF models perform the best overall, and in the focus months of January and July in particular, they are selected for further peak hour predictions and analyses.
For the peak day prediction, RF models also perform better than  or similar to GBM models, and the LR models do not converge due to their inability to handle imbalanced data sets.
Thus, RF models are also selected for peak day prediction to be consistent with the model choice for peak hour prediction.


\subsubsection{Feature importance}
\begin{figure}[!t]
	\centering
	\begin{subfigure}[b]{\fwidth}
		\includegraphics[width=1\linewidth]{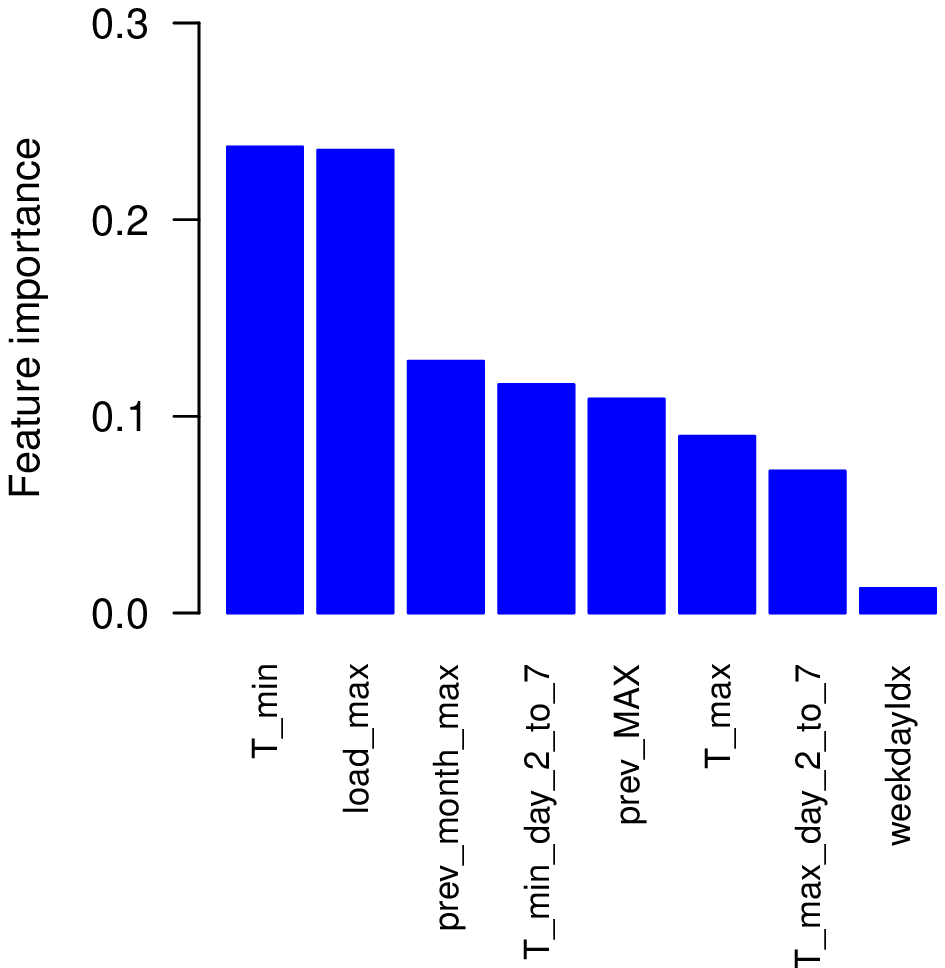}
		\caption{January}
	\end{subfigure}
	\begin{subfigure}[b]{\fwidth}
		\includegraphics[width=1\linewidth]{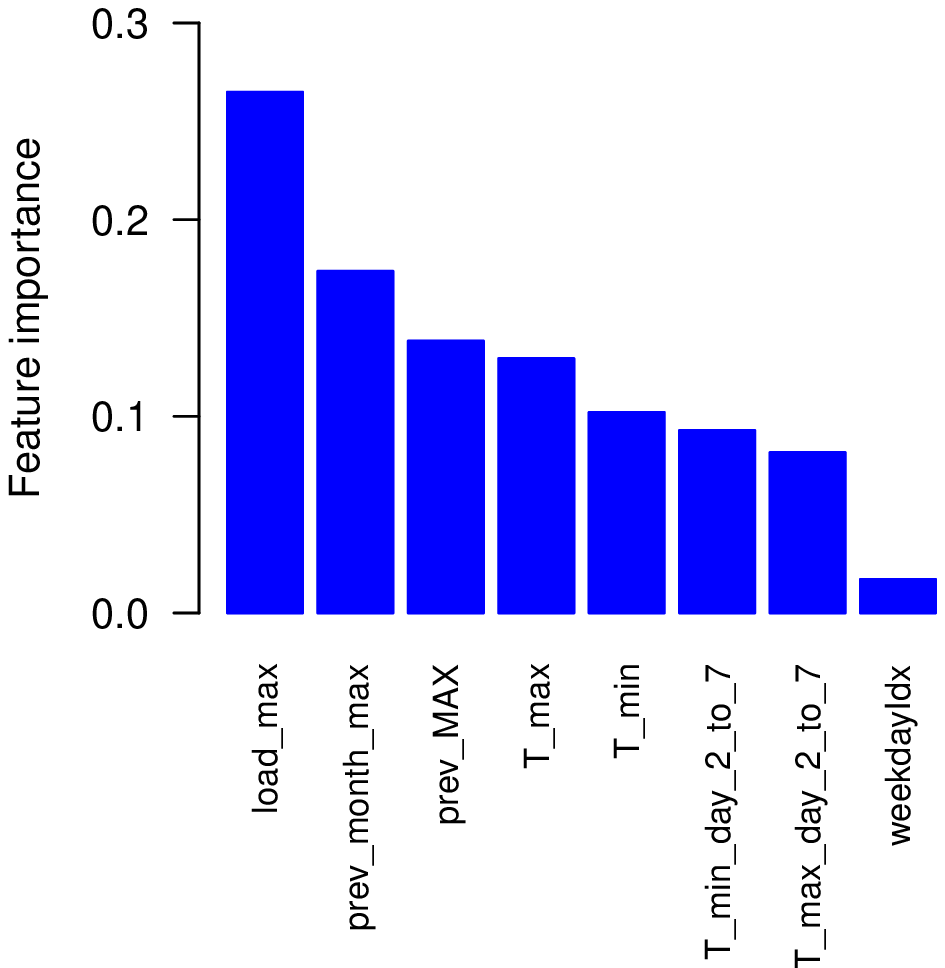}
		\caption{July}
	\end{subfigure}
	\caption{Feature importance of peak day models for January  and July.} \label{fig:FI_peak_day}
\end{figure}

Feature importance \citep{saarela2021comparison}, which can be directly calculated from the RF model, measures the relative importance of each predictor in the fitted model.
Fig.~\ref{fig:FI_peak_day} shows the ranked importance of the eight predictors in the January and July peak day models.
For the January model, \texttt{T\_min} (the minimum forecasted temperature) and \texttt{load\_max} (the maximum forecasted load for the operation day) are the most significant predictors, with almost identical factors and feature importance values of 0.237 and 0.235, respectively.
The other six variables are secondary.
For the July model, \texttt{load\_max} is more important than the other predictors.
Load-related predictors (ranking from 1 to 3) are slightly more important than temperature-related predictors (ranking from 4 to 7).

\begin{figure}[!t]
	\centering
	\begin{subfigure}[b]{\fwidth}
		\includegraphics[width=1\linewidth]{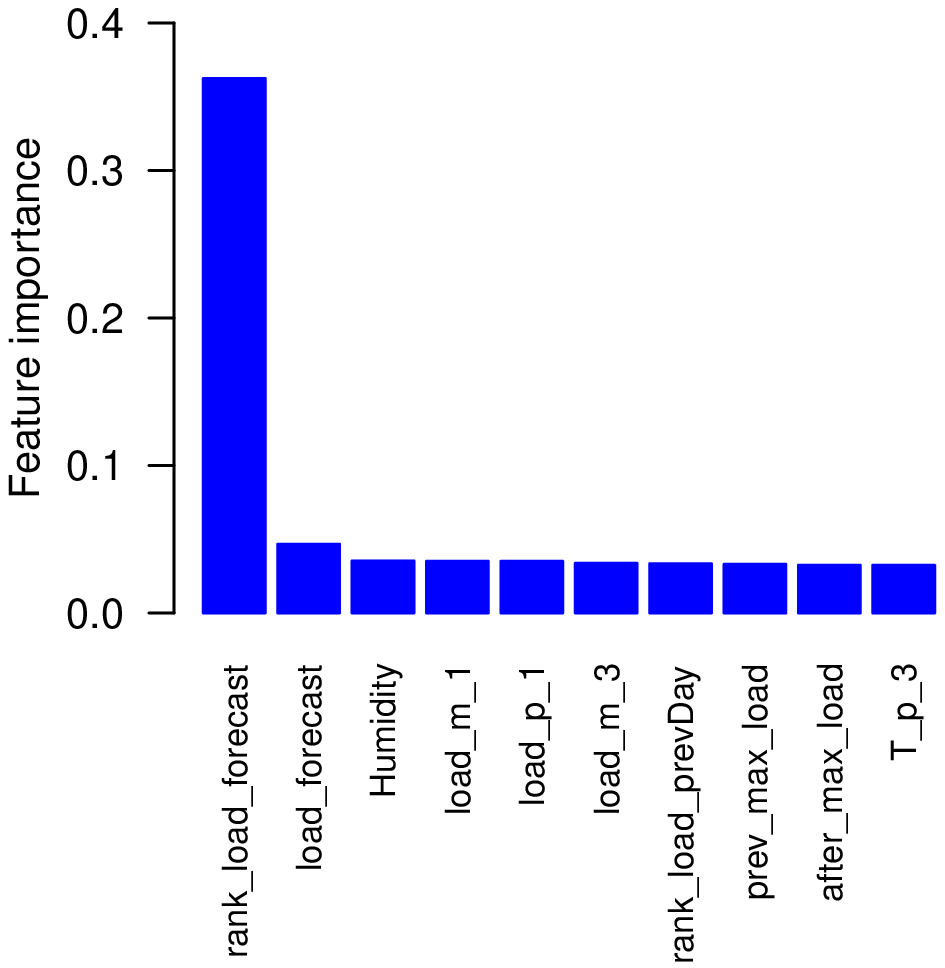}
		\caption{January}
	\end{subfigure}
	\begin{subfigure}[b]{\fwidth}
		\includegraphics[width=1\linewidth]{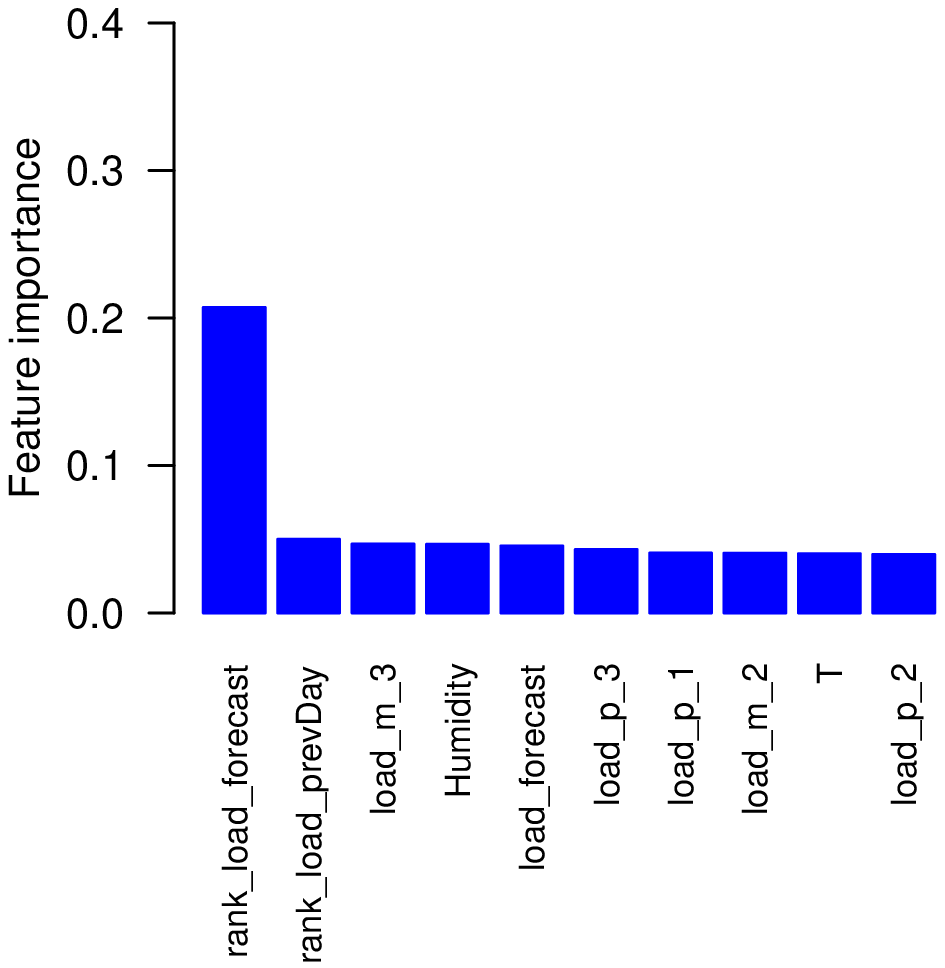}
		\caption{July}
	\end{subfigure}
	\caption{Feature importance of peak hour models for January and July.} 	\label{fig:FI_peak_hour}
\end{figure}

For peak hour predictions, Fig.~\ref{fig:FI_peak_hour} shows the top-ranked 10 of the 23 predictors with scaled importance in the January and July RF models.
\texttt{Rand\_load\_forecast} (the rank of the load forecast of each hour) is the dominant predictor for both models, and the feature importance values are 0.362 and 0.207, respectively.
Other import predictors include the load forecast for the January model and the rank load of the previous day for the July model.
To better understand the importance of predictors, one can refer to our principal component analysis and shrinkage discriminant analyses showing cross-dependence among predictors and response variables, as provided in Appendix A.

\subsection{Verification of peak day and peak hour predictions}
The above ML classification models yield parameter understanding and peak day/peak hour probabilities.
However, in order to use such outputs for a typical BESS dispatch decision-making, thresholds are needed for finalizing the peak days and peak hours, considering that 1) the maximum number of charging/discharging cycles should be less than 100 per year, and 2) the BESS can be charged/discharged up to 2 hours per day.
Given the first consideration, peak day prediction is set to be an exceedance probability (e.g., 3\%) as detailed below; and based on the second consideration, peak hour prediction is the top 2 hours with the highest peak probabilities in the operation day.

\subsubsection{Peak day prediction}
\begin{table}[!t]
	\caption{Comparison of BESS operation cycles and peak days captured between monthly peak day model and up-to-date peak day model.} \label{table-BESS_cycles_peak_days}
	\centering
	\begin{tabular}{lllll}
		\hline
		\multicolumn{1}{l}{\multirow{2}{*}{Year}} & \multicolumn{2}{l}{BESS operation cycles}                               & \multicolumn{2}{l}{Captured actual peak days}                           \\ \cline{2-5}
		\multicolumn{1}{l}{}                      & \multicolumn{1}{l}{Direct model} & \multicolumn{1}{l}{Indirect model} & \multicolumn{1}{l}{Direct model} & \multicolumn{1}{l}{Indirect model} \\
		\hline
		2001& 64& 78& 12& 12\\
		2006& 64& 63& 11& 12\\
		2008& 64& 72& 11& 12\\
		2011& 68& 71& 12& 12\\
		2019& 77& 87& 12& 10\\
		2020& 65& 83& 10& 11\\
		\hline
		Average& 67.2& 75.7& 11.3& 11.5\\
		\hline
	\end{tabular}
\end{table}

After the next operation day is assigned a peak probability by the RF model, a decision to charge/discharge a BESS on this day can be made by comparing the predicted peak day probability with a predefined threshold.
The BESS would only be operated if the predicted peak day probability is above the threshold.
A starting value of such a threshold is 3\%, with the assumption that any day within a month could be a peak day.
As shown in Table~\ref{table-BESS_cycles_peak_days}, both the direct and modified peak day models perform well in terms of the number of operation cycles and peak days captured per year for the 6 testing years.
The BESS would be operated for a similar number of cycles on average, 67 and 76, which are much fewer than the maximum 100 cycles/year requirement.
The average numbers of captured peak days are 11.3 and 11.5 per year, for the directed and modified peak day models, respectively.

\begin{figure}[!t]
	\centering
	\includegraphics[width=0.7\columnwidth]{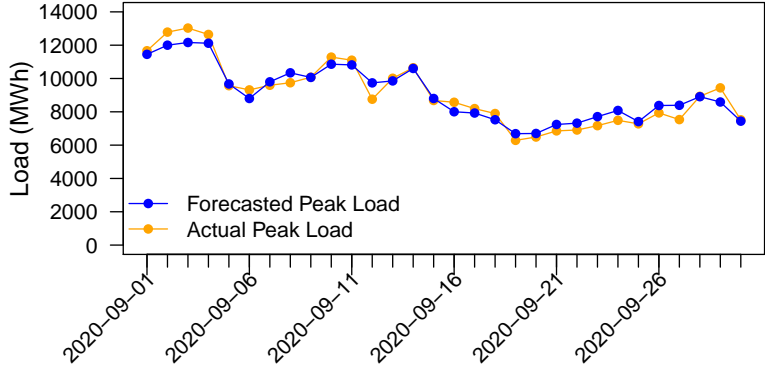}
	\caption{Forecast and actual daily peak load of September 2020.} \label{fig:Peak_load_September_2020}
\end{figure}

Table~\ref{table-actual_peak_day_probs} shows the predicted peak day probabilities for each month in 2020 using the direct and indirect peak day models.
With 3\% as the threshold, the direct peak day models miss the peak days in April and October, while the indirect peak day models only miss the peak day in September.
As shown in Fig.~\ref{fig:boxplot_peak_load}, compared with the historical peak loads, the maximum peak loads on April 8, 2020 (8324 MWh), and October 8, 2020 (8790 MWh), are at the lower end of the historical peak day loads, which causes the direct models to assign a less than 1\% probability to both  days and miss them.
In comparison, the indirect models can still capture these days with assigned probabilities of 3.8\% and 8.6\%, respectively.
For the peak day on September 3, 2020, the predicted peak day probability is only 2.9\% from the indirect peak day model, although its peak load of 13,027 MWh is higher than the median historical peak day loads.
The reason for such a low-probability estimate is because the neighboring days have comparable high load and the forecast peak load on September 3, 2020 (12,251 MWh), is lower than the actual peak load on September 2, 2020 (12,785 MWh), as shown in Fig.~\ref{fig:Peak_load_September_2020}, which shows the forecast and actual daily peak loads in September 2020.

\begin{table}[!t]
	\caption{Comparison of predicted probabilities of actual peak days in 2020 between using the direct and indirect peak day models.}
	\label{table-actual_peak_day_probs}
	\centering
	\begin{tabular}{llll}
		\hline
		\multicolumn{1}{l}{\begin{tabular}[l]{@{}l@{}}Actual\\ peak day \end{tabular}} & \multicolumn{1}{l}{\begin{tabular}[l]{@{}l@{}}Maximum hourly\\ peak load (MWh) \end{tabular}} & \multicolumn{1}{l}{\begin{tabular}[l]{@{}l@{}}Direct peak\\ day model \end{tabular}} & \multicolumn{1}{l}{\begin{tabular}[l]{@{}l@{}}Indirect peak\\ day model \end{tabular}} \\
		\hline		
		1/22& 12260& 13.0\%& 25.7\%\\
		2/22& 12065& 24.9\%& 27.1\%\\
		3/1&  10194& 11.7\%& 4.5\%\\
		4/8&  8324&  0.5\%& 3.8\%\\
		5/30& 9410& 8.7\%& 37.5\%\\
		6/29& 11606& 20.1\%& 28.0\%\\
		7/20& 13244& 27.1\%& 33.7\%\\
		8/27& 12511& 21.2\%& 30.3\%\\
		9/3& 13027& 11.9\%& 2.9\%\\
		10/8& 8790& 0.8\%& 8.6\%\\
		11/19& 10039& 35.4\%& 66.8\%\\
		12/9& 11625& 20.5\%& 16.4\%\\
		\hline
	\end{tabular}
\end{table}

\subsubsection{Peak hour prediction}
Practically, the BESS can be discharged for 2 hours during a peak day; therefore, the 2 hours with the highest probabilities from the peak hour model prediction in each day are selected for discharging.
The peak hour prediction performance is verified in two ways:
(1) comparing predictions from RF models with the naive model predictions using day-ahead load forecast;
and (2) evaluating the number of captured peak hours following the proposed peak day identification (e.g., 3\% exceedance probability) and peak hour prediction procedure.

\begin{figure}[!t]
	\centering
	\includegraphics[width=0.7\columnwidth]{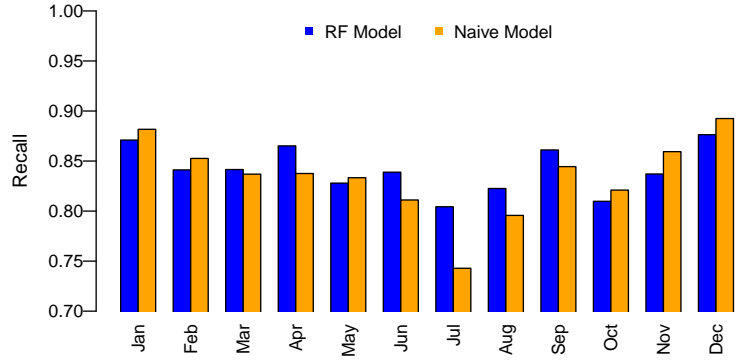}
	\caption{Comparison of prediction recall for each month between RF models and naive models.} \label{fig:prediction_precision_RF_vs_EIA}
\end{figure}
Fig.~\ref{fig:prediction_precision_RF_vs_EIA} shows the percentage of peak hours captured (true positives) relative to the total number of true positives and false negatives, $TP/(TP+FN)$, in each month for the 6 testing years.
The percentage metric, i.e., recall, is a good measure of classification model performance for imbalanced data like in this study.
For both RF and naive models, the recall values are highest in December and January and lowest in July and August.
The RF model has better overall performance than the naive model: of the 72 peak hours in the testing data, the RF model captures 67 (93.1\%) while the naive model captures 63 (87.5\%).
In particular, the recall of the RF model for July is two times higher than in the naive model.

Further evaluation of the entire prediction framework performance is based on the number of peak days and peak hours captured for all 6 testing years, using a 3\% exceedance probability threshold for peak day prediction and a 2-hour battery discharging time.
The results are summarized in Table~\ref{table-BESS_performance}.
The average number of peak hours captured is 10.8 per year for the 6 testing years.
All 12 monthly peak hours in 2006 and 2008 are captured.
For testing years 2011 and 2020, 11 of 12 peak hours are captured and the monthly peak hours in May 2011 and September 2020 are missed.
In 2001, peak hours in May and August are missed.
In 2019, peak hours in February, May, and September are missed.
The number of BESS charging/discharging cycles is 76 per year on average, ranging from 63 to 87, which meets the 100 cycles or less per year requirement.


\begin{table}[!t]
	\caption{Annual performance of BESS for 6 testing years using the proposed peak day and peak hour models.}
	\label{table-BESS_performance}
	\centering
	\begin{tabular}{llll}
		\hline
		\multicolumn{1}{l}{Year} & \multicolumn{1}{l}{\begin{tabular}[l]{@{}l@{}}BESS operation \\ cycles\end{tabular}} & \multicolumn{1}{l}{\begin{tabular}[l]{@{}l@{}}Number of peak \\days captured \end{tabular}} & \multicolumn{1}{c}{\begin{tabular}[l]{@{}l@{}}Number of peak \\ hours captured\end{tabular}} \\
		\hline		
		2001& 78& 12& 10\\
		2006& 63& 12& 12\\
		2008& 72& 12& 12\\
		2011& 71& 12& 11\\
		2019& 87& 10& 9\\
		2020& 83& 11& 11\\
		\hline
		Average& 76& 11.5& 10.8\\
		\hline
	\end{tabular}
\end{table}

\section{Conclusion}
In this paper, we presented an advanced ensemble ML framework for predicting peak days and peak hours to better support BESS dispatch decision-making.
The proposed approach features the probabilistic definition of peak day and peak hour,
a comprehensive set of physical and temporal factors and predictors,
nonlinear ensemble ML model implementation and selection, and data augmentation.
With cross-validation and model comparison, the proposed ML framework has been proven to work effectively.
The study also provided guidance on the model choices and favorable conditions for applying the proposed approach.
The study generated an ML-enabled dataset including cleaned historical data and derived attributes as exploratory and response variables.
One area of future work is to develop methods to explicitly integrate mid- to long-term factors into forecast models.
Another interesting research direction is to fully integrate state-of-the-art ML techniques such as ANN and LSTM to improve the efficiency and accuracy of peak predictions.

\addcontentsline{toc}{section}{References}
\renewcommand{\baselinestretch}{1}
\bibliographystyle{model5-names}
\biboptions{longnamesfirst,semicolon}
\Urlmuskip=0mu plus 1mu\relax

\appendix

\section{Exploratory data analysis of predictor matrix}
\subsection{Principal component analysis (PCA)}
\begin{figure}[!t]
	\centering
	\begin{subfigure}{0.65\columnwidth}
		\includegraphics[width=0.95\textwidth]{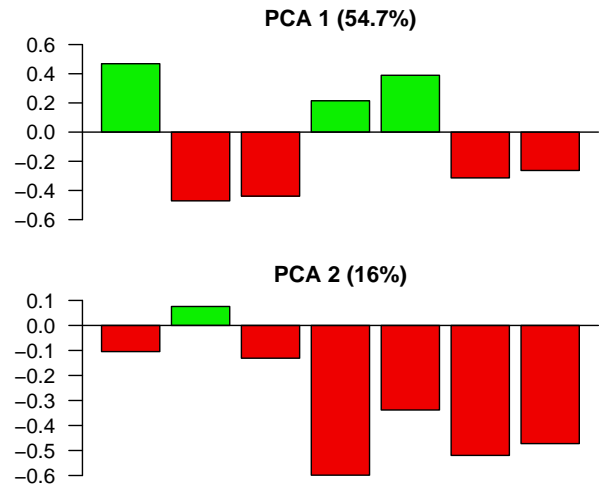}
		\caption{January}\label{fig:PCA_peak_day_Jan}
	\end{subfigure}
	\begin{subfigure}{0.65\columnwidth}
		\includegraphics[width=0.95\textwidth]{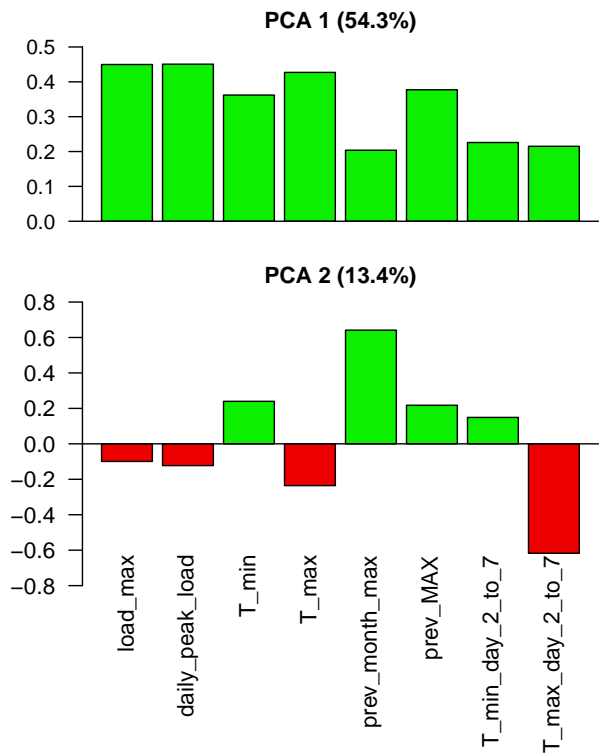}
		\caption{July}
	\end{subfigure}
	\caption{PCA loading plots of peak day predictors and daily peak load.} 	\label{fig:PCA_peak_day}
\end{figure}

PCA is performed to quantify cross dependence among predictors and the daily peak load (\texttt{daily\_peak\_load}). 
As examples, the loading plots for January and July are provided in Fig.~\ref{fig:PCA_peak_day}.
As can be seen, for the January peak day model, the first two components can explain about 71\% of the total system variance.
The maximum actual load, the forecast maximum load, the minimum temperature, and the maximum temperature of the operation day contribute the most to the first PC, PC1.
As expected, the maximum actual load has a strong positive correlation with the forecast maximum load and strong negative correlations with both the minimum and maximum temperatures.
For the July peak day model, the same four variables contribute the most to PC1: the maximum actual load, the forecast maximum load, the minimum temperature, and the maximum temperature of the operation day.
Positive correlations exist between the maximum actual load and both the minimum and maximum temperatures for the operation day.
The correlation between the maximum actual load and the maximum load on the previous day is stronger in July than in January.

\begin{figure}[!t]
	\centering
	\begin{subfigure}{0.65\columnwidth}
		\includegraphics[width=0.95\textwidth]{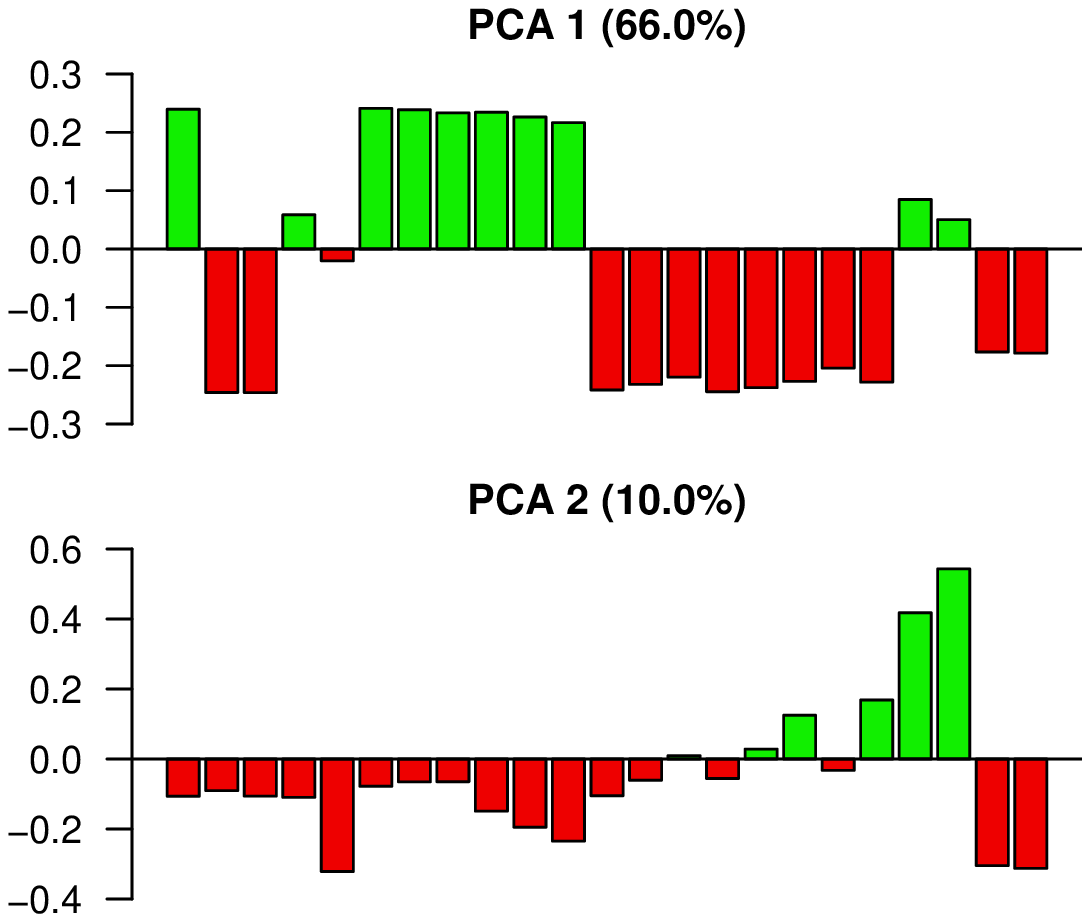}
		\caption{January}
	\end{subfigure}
	\begin{subfigure}{0.65\columnwidth}
		\includegraphics[width=0.95\textwidth]{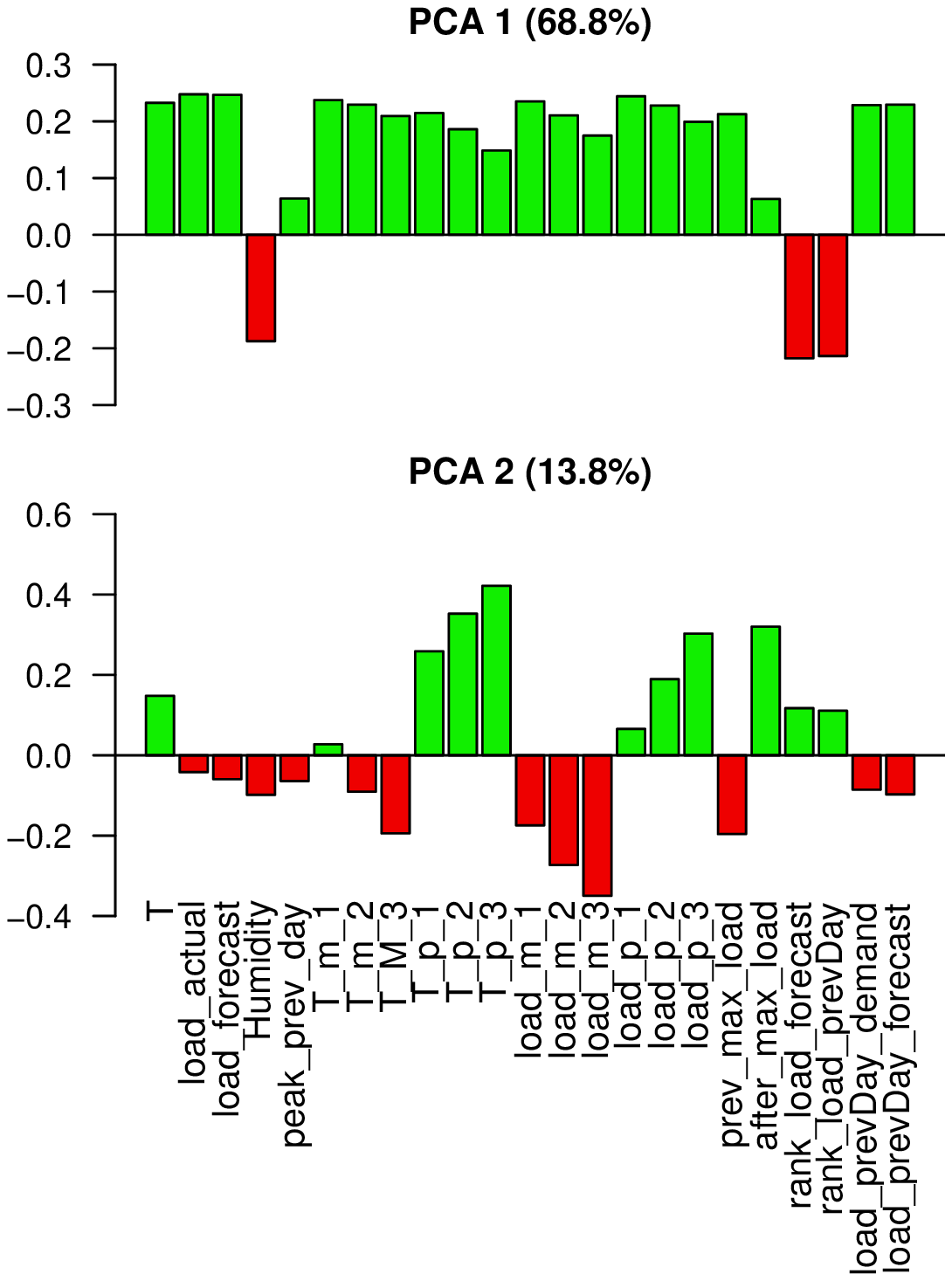}
		\caption{July}
	\end{subfigure}
	
	\caption{PCA loading plots of peak hour model predictors for January and July.} \label{fig:PCA_peak_hour}
\end{figure}

PCA plots for finer hourly resolution data are provided in Fig.~\ref{fig:PCA_peak_hour}, which shows that the first two components explain about 76\% and 83\% of the total variance for January and July, respectively.
Similar to the peak day models, negative correlations between load-related variables and temperature-related variables are observed in January, and positive correlations are observed in July.

\subsection{Discriminant analysis}
The shrinkage discriminant analyses (SDA) are performed to evaluate the ranking of the parameters for both the peak day and  peak hour models.
The analysis results are summarized in Table~\ref{table-sda}.
The peak day SDA shows that the forecast maximum load is the dominant predictor for both the winter and summer peak day models.
During the winter season, the forecast minimum temperature for the operation day is the second most important predictor, while in the summer season, the forecast maximum temperature for the operation day becomes the second most important predictor.
In both the January and July peak hour models, the rank of the load forecast for the predicting hour is the most important predictor.
Other important predictors include the temperature at the operation hour, and whether the predicting hour is the peak hour on the previous day in both models.

\begin{table}[!t]
	\caption{Top ranked predictors for the peak day and peak hour models from SDA.}\label{table-sda}
	\centering
	\resizebox{\textwidth}{!}{
		\begin{tabular}{lllll}
			\hline
			\multicolumn{1}{l}{\multirow{2}{*}{Rank}} & \multicolumn{2}{l}{Peak day model}                               & \multicolumn{2}{l}{Peak hour model}                           \\ \cline{2-5}
			\multicolumn{1}{l}{}                      & \multicolumn{1}{l}{January} & \multicolumn{1}{l}{July} & \multicolumn{1}{l}{January} & \multicolumn{1}{l}{July} \\
			\hline
			1& \texttt{load\_max}& \texttt{load\_max}& \texttt{rank\_load\_forecast}& \texttt{rank\_load\_forecast}\\
			2& \texttt{T\_min}& \texttt{T\_max}& \texttt{load\_forecast}& \texttt{T}\\
			3& \texttt{prev\_MAX}& \texttt{weekdayIdx}& \texttt{T}& \texttt{peak\_prev\_day}\\
			4& \texttt{T\_max\_day\_2\_to\_7}& \texttt{T\_min}& \texttt{peak\_prev\_day}& \texttt{T\_p\_1}\\
			5& \texttt{prev\_month\_max}& \texttt{prev\_MAX}& \texttt{rank\_load\_prevDay}& \texttt{load\_p\_3}\\
			\hline
	\end{tabular}}
\end{table}
\clearpage
\bibliography{IEEEfull,dwFull,DWpapers,peak_day_peak_hour}
\end{document}